\documentclass[conference]{IEEEtran}
\IEEEoverridecommandlockouts
\usepackage{cite}
\usepackage{amsmath,amssymb,amsfonts}
\usepackage{algorithmic}
\usepackage{graphicx}
\usepackage{textcomp}
\usepackage{xcolor}
\usepackage{multirow}
\def\BibTeX{{\rm B\kern-.05em{\sc i\kern-.025em b}\kern-.08em
    T\kern-.1667em\lower.7ex\hbox{E}\kern-.125emX}}
\begin{document}

\title{FIGO: Fingerprint Identification Approach Using GAN and One Shot Learning Techniques\\

}

\author{\IEEEauthorblockN{Ibrahim Yilmaz$^\S$ and Mahmoud Abouyoussef$^\dagger$}

\IEEEauthorblockA{\textit{$^\S$Department of Computer Science, Tennessee Technological University, Cookeville, TN, USA}\\
\textit{$^\dagger$Department of Computer Science and Engineering, University of Central Arkansas, Conway, AR, USA}}

Emails: iyilmaz42@tntech.edu and mabouyoussef@uca.edu}

\IEEEoverridecommandlockouts
\IEEEpubid{\makebox[\columnwidth]{979-8-3503-3698-6/23/\$31.00~\copyright2023 IEEE\hfill} \hspace{\columnsep}\makebox[\columnwidth]{ }}

\maketitle

\IEEEpubidadjcol

\begin{abstract}
Fingerprint evidence plays an important role in criminal investigations for the identification of individuals. The performance of traditional \textit{Automatic Fingerprint Identification System} (AFIS) depends on the presence of valid minutiae points and still requires human expert assistance in feature extraction and identification stages. Based on this motivation, we propose a Fingerprint Identification approach based on a Generative adversarial network and One-shot learning techniques (FIGO). Our solution contains two components: (a) fingerprint enhancement tier and (b) fingerprint identification tier. First, we propose a Pix2Pix model to transform low-quality fingerprint images into higher levels of fingerprint images. Furthermore, we develop another existing solution based on Gabor filters as a benchmark to compare with the proposed model by observing the fingerprint device's recognition accuracy. Experimental results show that our proposed Pix2pix model has better support than the baseline approach for fingerprint identification. Second, we construct a fully automated fingerprint feature extraction model using a one-shot learning approach to differentiate each fingerprint from the others in the fingerprint identification tier. Using the proposed method, it is possible to learn necessary information from one training sample with high accuracy. 

\end{abstract}


\section{Introduction}\label{sec:introduction}

The presumption of innocence is a key legal principle in democratic nations, where all individuals are considered innocent until proven guilty. This principle emphasizes the importance of providing clear evidence to the competent authorities in order to accurately detect crimes and identify criminals, ensuring a fair decision-making process.

Eyewitness identification has been historically used to uncover information about a crime, but it is an unreliable and problematic technique due to the error-prone nature of human memory and the potential for intentional deception. As a result, experts in the criminal justice system have turned to alternative solutions for criminal identification, including biometric technology. Biometric information is more reliable and accurate since it is directly related to an individual's unique characteristics, such as their voice, fingerprint, or iris. 

Law enforcement agencies are increasingly using state-of-the-art biometric technologies for the identification, authentication, and investigation of individuals. However, developing a reliable biometric system poses challenges and limitations. Facial identification, for instance, is influenced by various factors, including age, expression, and observation angle, which can compromise its accuracy. Iris recognition technology requires advanced visual sensing cameras to capture the necessary features with sufficient details. Fingerprint-based criminal detection systems, on the other hand, are widely used because of their unique and unchanging patterns \cite{awad2012machine}. 

    The performance of the AFIS depends on the extraction of the quality of relevant features. These features mainly consist of \textit{minutiae points} that are a set of points representing the distinctive character of a person and remain constant throughout a person’s lifetime (i.e., ridge ending, ridge, dot, etc.) \cite{kaur2008fingerprint}. The analysis, comparison, evaluation, and verification (ACE-V) framework is used by examiners to extract minutiae points from a latent fingerprint image (collected from a crime scene) and these points are side-by-side compared with the points taken from reference registered fingers \cite{abraham2013modern}. The system automatically identifies the person if any of the two sets of point pairs sufficiently mated as a result of these comparisons. Nevertheless, this system does not have a standard code of conduct for identification conclusions. For example, some examiners set threshold values as between 7 and 12 for forensic identification \cite{busey2021characterizing}. These differences in threshold values might lead to different conclusions. Another drawback of this method is the lack of a standard for the definition of the region of interest for the fingerprint image and as a result, different regions of interest size can contribute to different outcomes.

 \begin{figure}
\centering
\includegraphics[scale = 0.43]{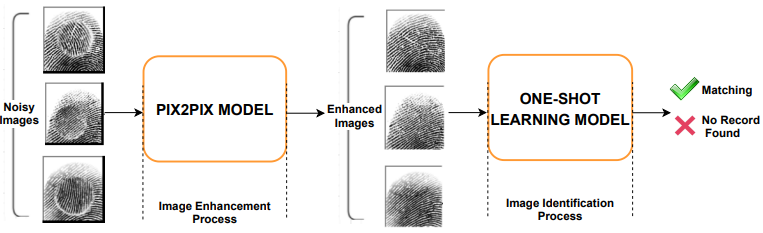}
\caption{Overview of the proposed fingerprint identification approach.      
}
\label{FIGO}
\end{figure}

To eliminate the need for manual intervention by human experts in traditional methods for fingerprint recognition, machine learning techniques can be used to automatically learn implicit patterns in fingerprint images. However, machine learning models require large training datasets \cite{9771790}, which is challenging in fingerprint recognition due to the uniqueness of each person's fingerprint. Traditional machine learning models also struggle to learn complex features from a single supervised sample and require retraining when new data is incorporated. To address these limitations, the one-shot learning approach is proposed, where the model is trained with a single sample. This approach turns the classification problem into a difference-evaluation problem, where the model learns similarity functions for image pairs instead of relying on manually designed features. As a result, the model can learn from a single example and eliminate the need for retraining in a dynamic environment, increasing the effectiveness and usability of criminal record applications.

Moreover, latent fingerprints collected from a crime scene are often not ideal for identification, being corrupt or incomplete. A pre-processing step to enhance the fingerprint information is therefore essential for the model to produce reliable results. Previous studies have used Gabor filter algorithms \cite{wang2008design}, but these generate artificial features that impede the success of fingerprint identification systems. To improve the identification process, the authors introduce a Pix2Pix model based on a conditional Generative Adversarial Network (cGAN) \cite{mirza2014conditional} that captures the relationship between original fingerprint images and their reconstructed versions. This enables the automatic reconstruction of distorted or incomplete images.

This paper proposes a Fingerprint Identification approach based on a Generative adversarial network and One-shot learning technique (FIGO). The motivation of this paper, as illustrated in Figure \ref{FIGO}, is to propose a sequential multi-model system where we synergistically bring two models together for improving image reconstruction and identification accuracy. Our research has the following contributions:

\begin{itemize}

    \item \textit{To the best of our knowledge}, this paper presents the first fingerprint identification model that uses a one-shot learning approach for fully automatic feature extraction and classification without human intervention. While our proposed model addresses the limitations of previous research on fingerprint classification, it is susceptible to significant performance degradation when low-resolution fingerprints are inputted into the system, as expected.
    
    \item In order to improve the quality of various levels of noisy fingerprint images, we develop a Pix2Pix model structure. Experimental results indicate that our proposed technique is better suited to reconstructing fingerprint images when compared with a conventional method (Gabor filter approach) and has shown substantial improvements in fingerprint classification accuracy when incorporated into the fingerprint verification system.  

    \item  We propose a novel approach that combines generic and automatic classification methods to handle a large real-world fingerprint dataset. This method enables effective learning from a single example in noisy environments, which is essential for accurate identification purposes.
    
\end{itemize}

The rest of this paper is organized as follows: the literature review in the context of our work is discussed in Section \ref{related}. The background information relevant to our study is reviewed in Section \ref{background}. The proposed techniques are presented in Section \ref{core}. Section \ref{implementation} describes the implementation of the reconstruction fingerprint and fingerprint classification approaches. We evaluate the models' performances in Section \ref{results}. In conclusion, we finalize the paper in Section \ref{conclusion}.

\section{Related works}
\label{related}

Numerous researchers have studied the techniques of fingerprint identification and feature extraction. Wang et al. \cite{wang2009fingerprint} proposed a fingerprint classification method that uses a continuous orientation field and singular points as discriminative features. The singular points, such as core or delta points, were extracted using a modified Poincare index. The experiment was conducted on the FVC $2002$ \cite{FVC2002} and FVC $2004$ \cite{FVC2004} databases, and the overall classification accuracy was found to be $96.1$\%. Maio et al. \cite{maio1997direct} presented an automatic minutiae algorithm that efficiently detected minutiae points from the NIST fingerprint database \cite{watson1992nist}. The authors demonstrated that their proposed method outperformed other solutions based on the image binarization approach in terms of efficiency and robustness. Minutiae-based fingerprint recognition systems have since become increasingly popular due to the uniqueness of the minutiae. However, these systems fail to identify the target class when singularity or minutiae points do not exist in a fingerprint image due to noise or partial fingerprint availability.

In order to eliminate the limitation of minutiae-based algorithms in fingerprint recognition systems, machine learning-based solutions have been proposed by some other researchers. Militello et al. \cite {militello2021fingerprint} demonstrated the performance of pre-trained CNN architectures (AlexNet, GoogleNet, and ResNet) over two different fingerprint databases. A Bayesian classifier was implemented in \cite{leung2010improvement} while support vector machines are developed in \cite {alias2016fingerprint} for fingerprint classification. Li et al. \cite {li2010neural} introduced a fingerprint classification based on a deep learning approach to obtain global features. The optimum parameters of the neural network model were found through a genetic algorithm. Kouamo et al. \cite{kouamo2016fingerprint} proposed a model to identify the users based on their fingers using a deep neural network.  Ala et al. \cite{balti2013fingerprint} proposed a dimensionality reduction technique based on the Euclidean distance between the center point and their nearest neighbor bifurcation minutiae. By doing so, redundant and irrelevant features were eliminated. The remaining topmost features were then used to develop a fingerprint identification method based on a backpropagation neural network. Using the FVC $2002$ database, they evaluated their methods and found that their results were superior to those of several other methods. In the literature, the primary problem with implementing an efficient fingerprint application using machine learning models is the lack of training samples since each individual has only one unique object. The above-mentioned studies attempted to circumvent this issue in some way by utilizing data augmentation techniques \cite{yilmaz2021privacy}, \cite{yilmaz2021avoiding}. However, augmented data created from existing data are often noisy. In many cases, machine learning models struggle to distinguish between noisy and noise-free data, which reduces the performance of the model \cite{yilmaz2020improving}. Moreover, for each new sample, both state-of-the-art machine learning algorithms as well as transfer learning-based possible viable solutions are required to be retrained. Instead of synthesizing new fingerprint samples, we propose a one-shot learning approach to train the model with one training sample. Our proposed technique distinguishes fingerprints from each other more efficiently and eliminates the need for retraining the whole system. 

As we outlined previously, fingerprints are rarely of good quality. Due to the presence of noise, they are distorted and corrupted. Therefore, it is essential to incorporate a fingerprint enhancement technique into the fingerprint identification system to make the system robust and increase identification accuracy. Barnouti et al. \cite{barnouti2016fingerprint} proposed histogram equalization and compression methods in order to both increase the image quality and the processing speed. Using histogram equalization, the authors were able to increase the quality of the images while using principal component analysis to reduce the dimensions without losing much information in order to speed up the verification process. However, histogram equalization causes a change in the brightness of an image resulting in a considerable loss of image detail. Wang et al \cite{wang2004fingerprint} initially found an area that consists of a singular point and then offered to use a bandpass filter in the Fourier domain for image enhancement. It is, however, difficult to extract singular points from a poor-quality fingerprint image. 

\section{Background}

\label{background}

This section reviews basic concepts covered in our research.

\subsection {Overview of Fingerprint Recognition System :}

The typical fingerprint recognition system consists of two stages, enrollment, and identification \cite{jain2010fingerprint}. During enrollment, an individual's fingerprints are scanned and digitized by a biometric device. The resulting images are processed to extract unique features, which are then stored in a database. At the identification stage, when a fingerprint is presented, the same process is repeated to capture distinct features and compare them against the database of known individuals. If the similarity between the presented fingerprint and a stored record is deemed sufficient, the identification is deemed successful.



\subsection {Conditional Generative Adversarial Network :}

The Generative Adversarial Network (GAN) is a type of machine learning/deep learning model where two neural networks compete in a zero-sum game \cite{yilmaz2020addressing}. GANs consist of two networks, the \textit{Generator} and \textit{Discriminator}, where the generator produces candidates and the discriminator evaluates them. The generator learns to map from a latent space to a data distribution of interest, while the discriminator distinguishes the generator's candidates from the real data distribution. The goal is for the generator to create candidates that are indistinguishable from real samples, such that the discriminator is unable to tell the difference. In certain situations, the GAN may not generate the desired output because it lacks control over the generated data types. To address this limitation, the conditional Generative Adversarial Network (cGAN) was introduced in \cite{mirza2014conditional}. By using auxiliary information such as partial parts of the data, cGAN can guide the data generation process to produce better representations of target images.  

\section{The Proposed Approach}

The section is divided into three parts. First, we propose using the Pix2Pix model to enhance the pattern legibility of fingerprints, improving classification reliability. Second, we compare our approach with an existing Gabor filter solution by evaluating fingerprint identification accuracy. Finally, we discuss our unique one-shot learning method for fingerprint identification. Note that Pix2Pix and one-shot learning models are the two essential components of the FIGO model.    

\label{core}
\subsection{Fingerprint Reconstruction Method Using Pix2Pix Model:}

To address the issue of corrupted latent fingerprints due to external factors, we propose a new fingerprint reconstruction framework using the Pix2Pix model, inspired by \cite{isola2017image}. This model is based on conditional Generative Adversarial Networks (cGAN), where the network learns how to convert noisy and incomplete fingerprint images into enhanced versions through pixel-to-pixel translation. Two neural network models based on CNN, the discriminator (D) and generator (G), are used to learn the translation function automatically.

\begin{figure}
\centering
\includegraphics[width=5cm]{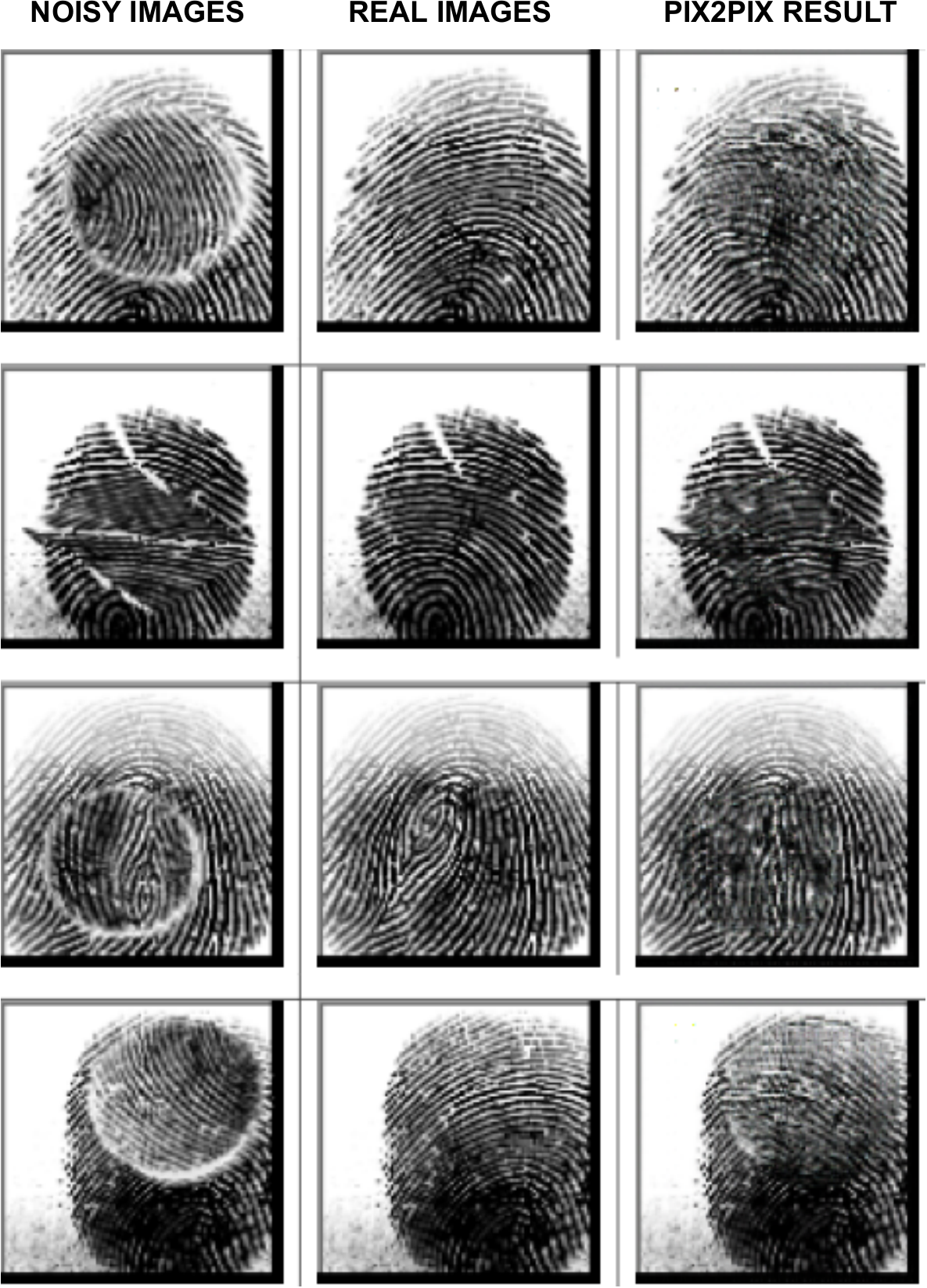}
\caption{Results of fingerprint image reconstruction using the Pix2Pix model.        
}
\label{cgan}
\end{figure}

The G and D models are trained simultaneously in the proposed fingerprint reconstruction framework. G takes low-resolution fingerprint images as input and produces an output, which is then sent to D along with distorted fingerprint images. D compares the two inputs and tries to detect G's fake samples. G attempts to fool D and translate distorted images into the desired restored images. By fine-tuning G parameters, the real and fake data distributions become similar, resulting in producing enhanced fingerprint images from the noisy ones. The success of the Pix2Pix model is shown in Figure \ref{cgan}. 

\subsection{Fingerprint Reconstruction Method Using Gabor Filters :}
\label{Gabor}

In \cite {gabor1946theory}, Dennis Gabor introduced the Gabor filter, which was later extended by David Daugman in \cite{daugman1980two}. The Gabor function is useful for capturing time and frequency information simultaneously from digital images due to its frequency-selective and orientation-selective properties. As a result, it has been widely used in computer vision and image processing for analyzing textures, feature extraction, and other applications.

Yang et al. \cite{yang2003modified} found that Gabor filters are not effective for improving fingerprint images that do not follow the ideal sinusoidal plane wave shape. Moreover, this method relies on a set of manually adjusted parameters, such as the standard deviations of the Gaussian function and the convolution mask size. However, determining the optimal parameter values through trial and error makes the method image-dependent.

To provide a more robust noise reduction algorithm while preserving the true ridge and valley structures without depending on fingerprint images, we propose a Pix2Pix model. Our experimental results in Section \ref{results} demonstrate that our Pix2pix model is more powerful than traditional Gabor filters in reconstructed fingerprint images.

\subsection{Fingerprint Identification Using One-Shot Learning:}

To develop a fully automated fingerprint classification system, we propose a similarity-based learning approach in which the model attempts to learn similarity functions for image pairs as opposed to using manually designed features. This idea is inspired by the Siamese network structure \cite{koch2015siamese}. The proposed model consists of two twin CNN networks connected by a similarity layer at the top. To put it another way, these two identical networks have the same configuration (i.e., the same number of layers and nodes) that share weights and parameters. The tying of the weights ensures that similar images will map close together in the feature space while dissimilar pairs will fall apart since each network in the architecture computes the same function.

\begin{figure}
\centering
\includegraphics[width=8cm]{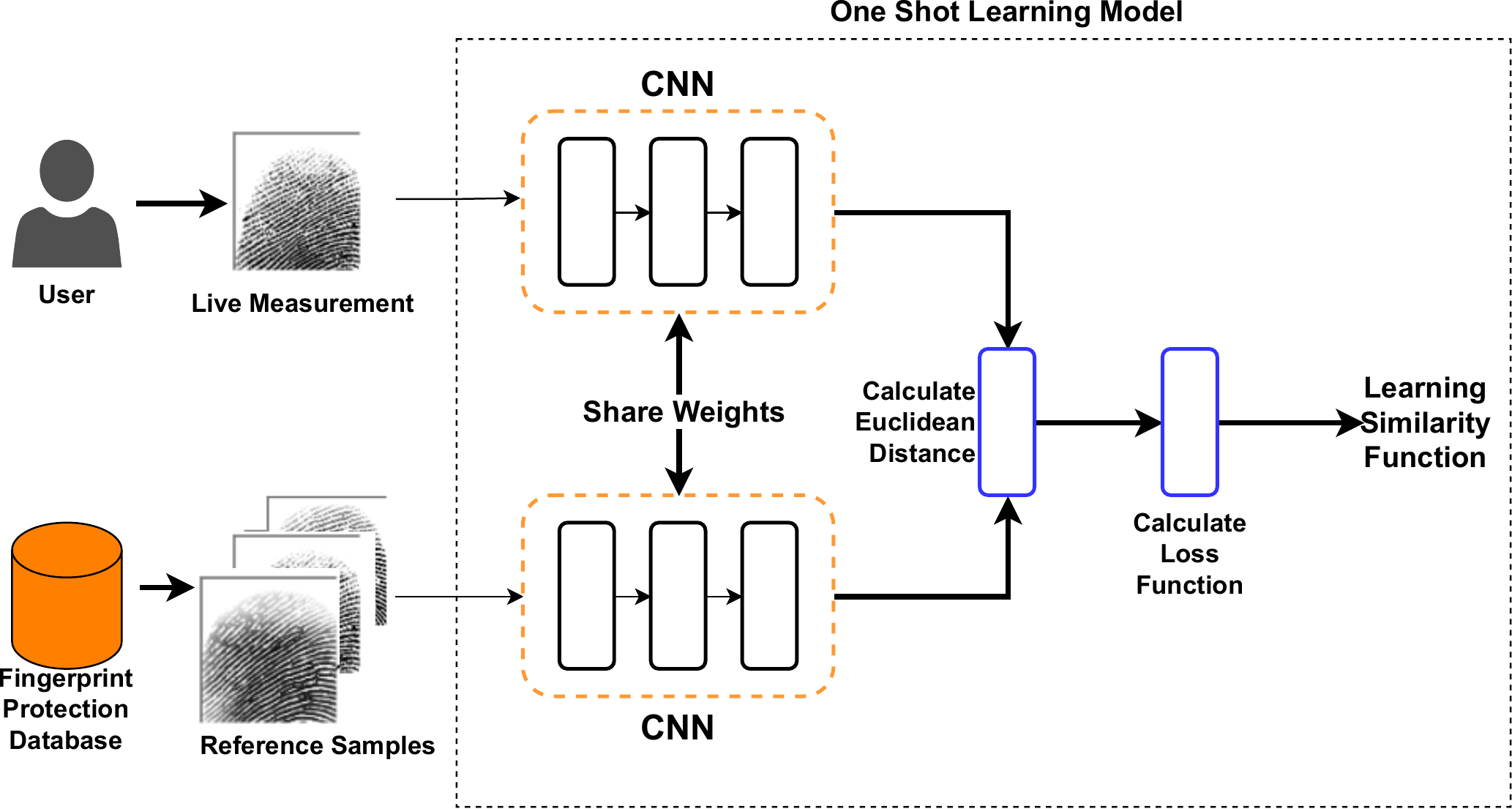}
\caption{Illustrative idea of the fingerprint classification using one-shot learning.      
}
\label{one_shot}
\end{figure}

To develop a model for fingerprint image classification, we generate all possible image pairs from original samples and label them as 'positive' (for the same images) or 'negative' (for different images). We then train CNN models to extract unique features from these labeled examples and calculate the Euclidean distance between all image pairs. Using this distance, we derive a loss function that enhances the model's discriminative and generalization ability. This allows the model to efficiently differentiate a fingerprint from others and accurately identify the person even with variations in the fingerprint image due to structured noise. The training process is depicted in Figure \ref{one_shot}.

Let M be the fingerprint identification model, x be any fingerprint image, and Y be the corresponding ground-truth output label (negative (0) or positive (1)) of each x * x pair. We define the cost function C(M, x * x, Y) to train the fingerprint classification model, as shown in the following equation. The goal is to minimize the error when a pair of inputs belong to the same class, while maximizing it otherwise. 

\begin{equation}
    objective \left\{\begin{array}{lll}
= minimize \rightarrow c(M, x1 * x2, Y), & \text { if } Y =1\\
= maximize \rightarrow c(M, x1 * x2, Y), & \text { if } Y =0\\
\end{array}\right.
\end{equation}

The distance metric is calculated based on element-wise absolute difference: 

\begin{equation}
   d\left(x_{1}, x_{2}\right)=\left|M\left(x_{1}\right)-M\left(x_{2}\right)\right| 
\end{equation}

The proposed fingerprint identification algorithm using the one-shot learning approach differs significantly from traditional machine learning techniques and offers two key advantages. Firstly, traditional machine learning models require a large number of training samples, while this model can be trained using just one sample. Secondly, there is no need to retrain the model when a new user is added or removed from the fingerprint database, since the model is trained to distinguish between fingerprints based on pairwise similarity. This is unlike typical machine learning models that require retraining with each new user, reducing their usability.

\section{Model Implementation}
\label{implementation}

In this section, we describe the dataset used to build the fingerprint identification model using generative adversarial networks and one-shot learning techniques, along with an explanation of how the FIGO model is implemented. 

\subsection{Dataset:}

In our project, we use the SOCOFing dataset in \cite{shehu2018sokoto}. The dataset contains $6,000$ fingerprints taken from $600$ Africans. A fingerprint sample is taken from each finger of each individual. The noise was created using three different alteration techniques, including obliteration, central rotation, and z-cut, using the STRANGE toolbox  \cite{papi2016generation}. Based on parameter settings in this toolbox, the authors generated different levels of noise and categorized them as '\textit{easy}',  '\textit{medium}', and '\textit{hard}'. The output of this phase is shown in Table \ref{tab:my_11} as an example. As seen in Table \ref{tab:my_11}, the original image is slightly distorted by noise in the 'easy' category. When the level of noise is increased in the 'medium' category, the quality of the image is reduced. In addition, the more additive noise corrupts the real image to a great extent in the 'hard' category. In this case, the fingerprint image will lose much of its most important information.       

\begin{table}[]
            \centering
            \caption{Illustration for alteration categories with the real sample}
            \begin{tabular}{|c|c||c|c|}
            \hline
              Data   & Instances & Data   & Instances \\ 
              \hline
                 Real & \includegraphics[width=0.05\textwidth]{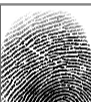} & Easy & \includegraphics[width=0.05\textwidth]{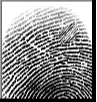}  \\ \hline
                   Medium & \includegraphics[width=0.05\textwidth]{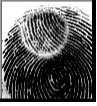}   &
                    Hard & \includegraphics[width=0.05\textwidth]{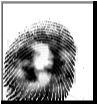}   \\ \hline
            \end{tabular}

            \label{tab:my_11}
        \end{table}

\subsection{FIGO implementation:}

The FIGO model, as previously stated, is comprised of a Pix2Pix model and a fingerprint identification model, both of which are trained separately during the training process. After training is completed, the models are linearly stacked into a pipeline, where the Pix2Pix model's output is passed to the proposed automated fingerprint identification model for identification. We implement both models in Python and TensorFlow. To train and evaluate the models, we randomly divide the SOCOFing dataset into three portions: training, testing, and verification, with 80\%, 10\%, and 10\%, respectively. The Python code is executed on Google Colab with the NVIDIA GPU Tesla P4 \cite{google}.

To develop a denoising framework based on the Pix2Pix model, we utilize the same architecture and parameters for both the discriminator and generator except for types of loss functions. A "Mean Absolute Error" is used to measure the error of the generator, whereas a "Mean Square Error" is used to measure the error of the discriminator. The layer size is $17$, learning rate is $0.0002$, the Epoch number is $10$, and the Optimizer is Adam Optimizer. 

In addition, we propose a novel one-shot learning approach for identifying criminals, utilizing only real, undistorted samples for training the model. This study uses embedding layer size of $2$, Similarity layer size of $6$, Learning rate of $0.001$, Epoch number of $10$, Adam optimizer, and Binary Cross Entropy loss function. During model training, we used triplet loss to create positive and negative pairs of data, minimizing the distance between positive samples and maximizing the distance between negative samples. This approach can ease the burden of generating or collecting large amounts of fingerprint data and effectively distinguish latent objects.

Once the FIGO model is trained, the model is tested on unseen data. During testing, the Pix2Pix model takes an unseen noisy fingerprint image as an input and produces an enhanced version. Each image from the real dataset is paired with this enhanced image one by one and fed into the fingerprint identification model. For each match, the fingerprint identification model computes a similarity score. The system determines the user's identity with the highest score.

\section{Evaluation and results}
\label{results}

\begin{figure}
\centering
\includegraphics[width=8cm]{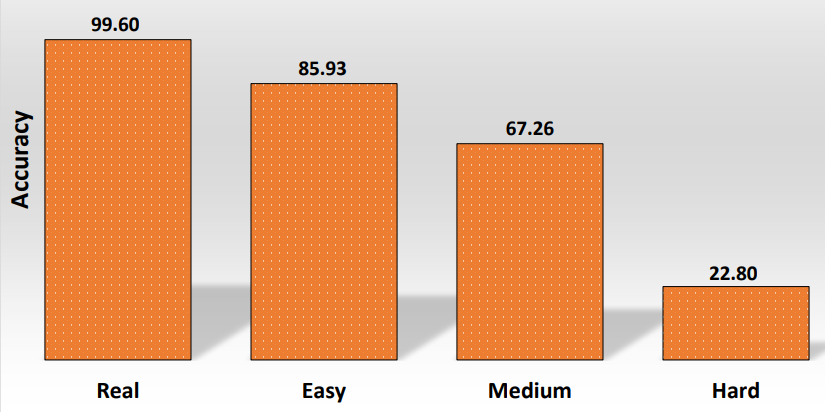}
\caption{The fingerprint identification model's recognizing accuracy under various corruption levels without fingerprint enhancement algorithms.}     
\label{One_shot_results}
\end{figure}

We conducted several experiments to assess the performance of the FIGO model in enhancing image quality and recognition accuracy. In the first set of experiments, we tested the model with real samples and corrupted samples under three different noise levels, without applying fingerprint enhancement algorithms as shown in Figure \ref{One_shot_results}. The model accurately classifies real fingerprint images, but its accuracy drops to $85.93$\% with minor alterations due to limited distinctive information. Severe degradation results in poor accuracy, and heavy noise contamination reduces identification performance to $22.8$\%.

\begin{figure}
\centering
\includegraphics[width=8cm]{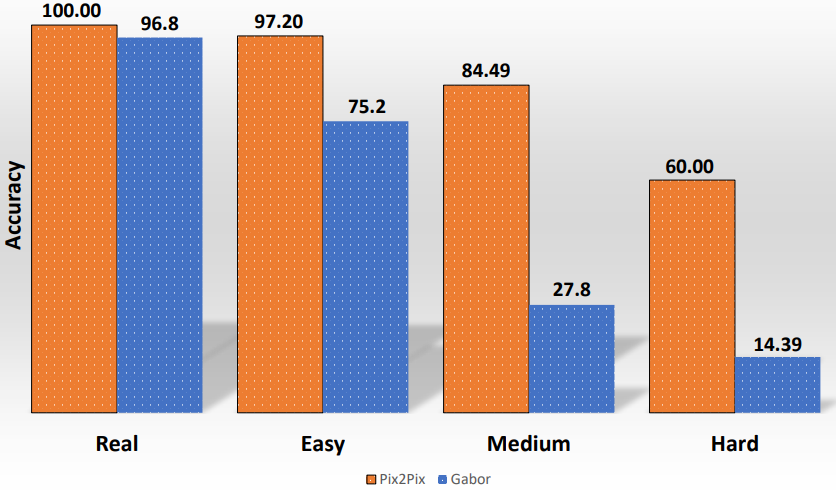}
\caption{The fingerprint identification model's recognizing accuracy under various corruption level by integrating fingerprint enhancement algorithms. }     

\label{gan_gabor_comp}
\end{figure}

Our second set of experiments evaluated our fingerprint verification model by integrating fingerprint enhancement algorithms (Pix2Pix model and Gabor filter) to improve recognition accuracy. The Pix2Pix model outperformed the Gabor filter method as shown in Figure \ref{gan_gabor_comp} due to its ability to extract more discriminative information from latent fingerprint images. However, the Gabor filter method generated spurious features that led to unsatisfactory accuracies. Our FIGO model (multimodal system with the Pix2Pix model) improved the recognition accuracy by approximately $12$\% and $17$\% for the easy and medium groups, respectively, due to its efficient noise reduction method. In severe cases of fingerprint image distortion, our FIGO approach improved the accuracy of fingerprint identification by approximately $40$\%.

 \begin{figure}
\centering
\includegraphics[width=8cm]{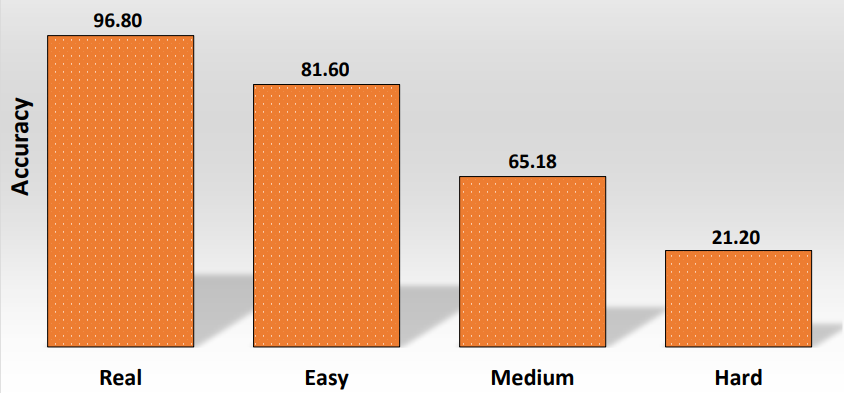}
\caption{The fingerprint identification model's recognizing accuracy under various corruption level by combining Gabor filter and Pix2Pix algorithms. }     
\label{gan_gabor_all}
\end{figure}

In our final set of experiments, we combined the Gabor filter method and Pix2Pix model to enhance fingerprints. The output of the Gabor filter technique was fed to the Pix2Pix model before the input images were entered into the fingerprint verification model. As shown in Figure \ref{gan_gabor_all}, the combined scenario slightly improved the matching performance of the fingerprint recognition system for noise-free images. However, in the presence of noise, the designed model performed worse than unimodal recognition, but still showed an improvement compared to Gabor-based enhancement. 

Overall, the results of the experiments indicate that the accuracy of the proposed fingerprint verification system based on the one-shot learning approach is very high with high-quality fingerprint images, but its performance decreases with distorted images. In our study, the Gabor filter method is not a desirable approach as an image enhancement process. Moreover, the proposed Pix2Pix model can handle different levels of noise and is more powerful than Gabor filter approaches for enhancing fingerprint images, so it can be incorporated into the identification process as a complementary tool.

\section{Conclusion}

The criminal justice system relies on forensic evidence to make decisions, but false or misleading evidence can lead to injustice and a breakdown of trust. Latent fingerprints are important evidence, but poor image quality and reliance on human expertise can affect accuracy and reliability. The FIGO model, a multimodal system combining the Pix2Pix model with fingerprint classification based on one-shot learning, ensures robust feature extraction and accurate matching without human involvement. Experiments showed that the proposed image enhancement algorithm significantly improves system performance and identification accuracy.
\label{conclusion} 

\bibliographystyle{IEEEtran}
\bibliography{main}


\end{document}